\documentclass{IOS-Book-Article}

\usepackage[T1]{fontenc}
\usepackage{mathptmx}
\usepackage{graphicx}
\usepackage{float}

\DeclareMathAlphabet{\mathcal}{OMS}{cmsy}{m}{n}

\def\hb{\hbox to 10.7 cm{}}

\begin{document}

\pagestyle{headings}
\def\thepage{}

\begin{frontmatter}              

\title{The Shape of a Benedictine Monastery: The \textsf{SaintGall} Ontology\\ (Extended Version)}


\author[A]{\fnms{Claudia Cantale,}}
\author[B]{\fnms{Domenico Cantone,}}
\author[C]{\fnms{Manuela Lupica Rinato,}}
\author[B]{\fnms{Marianna Nicolosi-Asmundo,}}
and
\author[B]{\fnms{Daniele Francesco Santamaria}}
\address[A]{Dept.\ of Humanities, University of Catania, Italy}
\address[B]{Dept.\ of Mathematics and Computer Science, University of Catania, Italy}
\address[C]{Officine Culturali, Catania, Italy}

\begin{abstract}
	We present an OWL 2 ontology representing the Saint Gall plan, one of the most ancient documents arrived intact to us, which describes the ideal model of a Benedictine monastic complex that inspired the design of many European monasteries. 
\end{abstract}

\end{frontmatter}

\section{Introduction}
Monasteries are conceived by the Benedictine monastic order, founded by Saint Benedict of Nursia, during the last period of the Western Roman Empire. The monastic shape aims preserving the European Christianity inside small self-sustaining communities where to lead a life of mystic and religious contemplation and introspection.\footnote{ \guillemotleft Monasticism has its root in the interpretation of the Christian faith developed in the theology of the VI century firstly in Orient. Analogously to theology and architecture, it is subjected to a deep transformation in Occident. [ ... ] The Benedictine Order remains for a long time the principal one. Hundreds of convents and monasteries are spread across the Christian Europe and represent cells of Christian tradition and faith, of science, and of culture\guillemotright\ \cite{Kuback2001}.}  The main principle is to protect and shield Christian religion and tradition from barbarian invasions.

Monasteries differ from convents primarily because of their purpose. Monasteries are inhabited by monks belonging to some monastic order such as the Benedictine one, having an ascetic and solitary lifestyle.
Convents, originated later with the mendicant orders, such as the Franciscan one, are more dependent by the outside world. The two religious constructions arise in different historical periods carrying out different functions inside the religious community. 
Starting from the VII century, Western Europe is characterized by a capillary network of monasteries. Their shape in Occident remained largely unchanged in its characteristics during  the whole Middle Age and in all Christian countries. 

Monasteries are often also abbeys that are spaces where the \textit{nullius diocesis} is effective. Such norm, in the canon law, represents the independence of a church and of the related monastery from the diocese in which  the building is located. Therefore, the abbot substitutes for the bishop inside the Benedictine ``village''.
 
\begin{figure} 	
 	\includegraphics[scale=0.5]{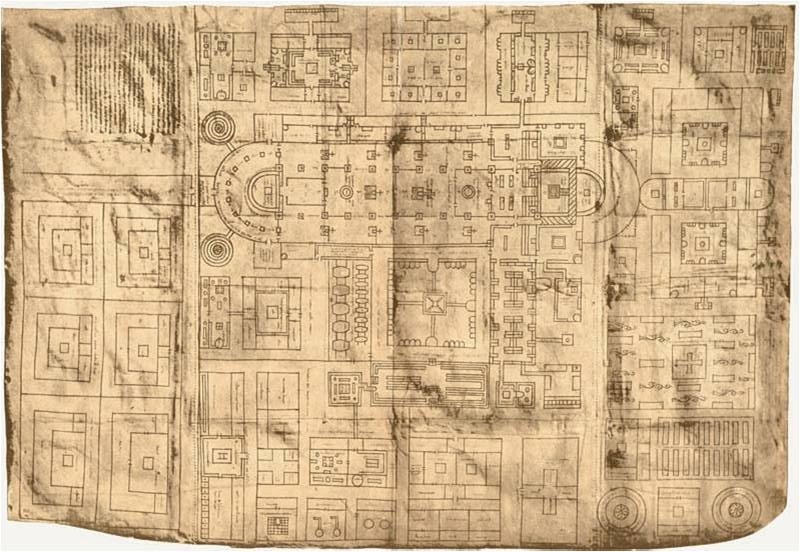}	
 	\caption{The plan of Saint Gall. }
 	\label{imgSantGall}
 \end{figure}

Strongly inspired by the rule of Saint Benedict, the plan of St.\ Gall, illustrated in Figure \ref{imgSantGall}, is a model of monastery  better representing the Benedictine architecture.\footnote{http://www.stgallplan.org/} Founded in the context of Pre-Romanesque Carolingian art and architecture, in which a varied 
partition of the space is preferred, it can be considered a fixed-type for the Middle Age monasteries \cite{DeVecchi95}. Moreover, being one of the most ancient  descriptions of primitive Benedictine monastery arrived intact today, it turns out to be an important  structural, architectonic, and functional landmark for the Benedictine monasteries. In the plan, St.\ Gall monastery is idealized together with its essential components. In fact, as it often happens in the context of architectural history, buildings realized in a long temporal window are subject to change with respect to the original idea because of historical, economical, practical, and morphological reasons. Many European monasteries are inspired by the St.\ Gall plan even if for practical and technical reasons they deviate from it. For instance,  Catania's Benedictine Monastery \cite{AtrCo08,DeCarlo88} contains most of the elements of the St.\ Gall plan with the exception of some locations such as the brewery that, for cultural reasons, is replaced by a distillery. Moreover, Catania's Benedictine Monastery is a urban monastery and therefore the structure of the animal farms is also slightly modified.
 
\smallskip
 In this paper we present an OWL 2 ontology, called \textsf{SaintGall} Ontology, representing the monastery described in the St.\ Gall plan. \textsf{SaintGall} Ontology has been developed by taking into account structural, architectonic,
and functional details of the buildings included in the plan, and information
provided by \cite{ DeVecchi95,Kuback2001,Gombrick1950,Willis}. It consists of more than 400 classes, almost 60 object properties, and more than 1000 logical axioms. It exploits OWL 2 constructs such as existential restriction and qualified cardinality restriction,  and has been classified using the Fact++ reasoner. 

\section{The Ontology of St.\ Gall plan}

The \textsf{SaintGall} Ontology\footnote{https://goo.gl/XN2hc3} describes buildings and green spaces depicted in the plan of Saint Gall considering their cardinal orientation, their position with respect to other entities inside the plan, and their architectonic, structural, and functional features.

The ontology exploits the following main classes. The class \texttt{Building} describes a generic building, \texttt{Garden} specifies a generic green space, \texttt{Element} describes
architectonic elements, natural elements, furnitures, spaces contained in the plan.
 The ontology also provides classes and properties to describe the cardinal orientation, position, and shape of the structures of the plan, and the role of people living inside the monastery. 

\begin{sloppypar}
At first we model the functional areas of the monastery classifying the buildings represented on the map according to their intended use. Specifically, 
we introduce as subclasses of \texttt{Building} the pairwise disjoint classes \texttt{BuildingForEducation}, \texttt{BuildingForHospitality},
\texttt{BuildingForTheSickAndInfirm}, \texttt{FarmBuilding},  \texttt{PrincipalMonasticBuilding}.  

\begin{figure} 
	\includegraphics[scale=.95]{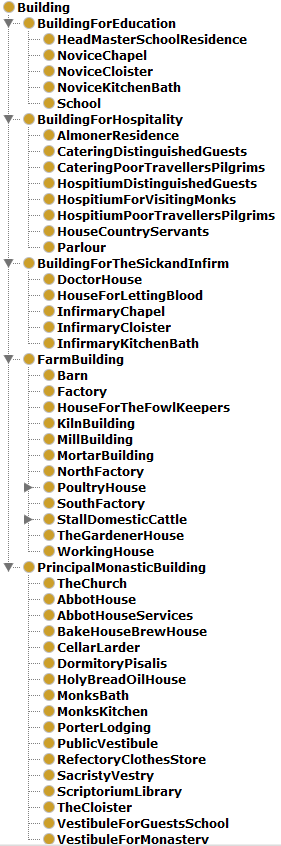}
	\caption{Subclass hierarchy of \texttt{Building}.}
   \label{buildimg}
 \end{figure}

\texttt{BuildingForEducation} includes, in particular, the class \texttt{School}, modeling a building intended for the education of scholars,  and the class \texttt{NoviceCloister}, representing the novice cloister, dwelling of young people oriented to the monastic life. \texttt{BuildingForHospitality} contains among others the class \texttt{HospitiumDistingueshedGuests}, modeling the hospitium for the reception of eminent strangers, and the class \texttt{HospitiumPoorTravelersPilgrims}, representing the dwelling of poor travelers and pilgrims. The class \texttt{BuildingForTheSickAndInfirm} contains in particular the subclass \texttt{InfirmaryCloister}, representing the cloister where the sick brethren are lodged, and the class \texttt{DoctorHouse}, containing among others a private room for the physician and a room for very ill patients. 
The class \texttt{FarmBuilding} models the factory, the working house, and other buildings devoted to domestic cattle, poultry, and their keepers.  The class \texttt{PrincipalMonasticBuilding} includes in particular the classes \texttt{AbbotHouse}, modeling the dwelling of the abbot, \texttt{TheCloister}, describing the cloister where monks live, and \texttt{TheChurch}, describing the abbey. The hierarchy of \texttt{Building} is shown in Figure \ref{buildimg}.

\begin{figure}	
\includegraphics[scale=0.3]{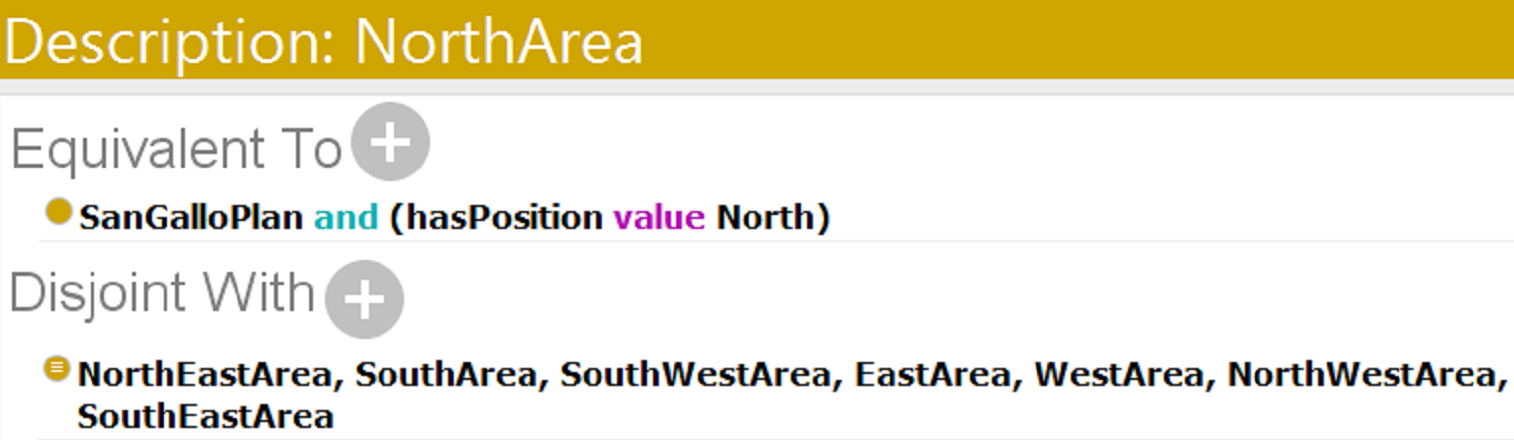}	
\caption{Description of \texttt{NorthArea}. }
\label{imgNorthArea}
\end{figure}

\begin{figure}	
	\includegraphics[scale=0.6]{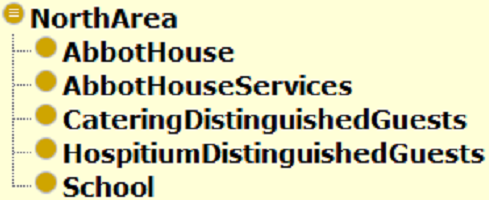}	
	\caption{Inferred hierarchy of \texttt{NorthArea}. }
	\label{imgNorthAreaInf}
\end{figure}

The green spaces inside the monastery are modeled by means of the class \texttt{Garden}, having the disjoint subclasses \texttt{Cemetery}, \texttt{KitchenGarden}, and \texttt{PhysicGarden}. Cardinal orientation of buildings and gardens on the map are modeled by the classes \texttt{CardinalDirection}, \texttt{CentralPosition}, and the object-property \texttt{hasPosition}, having as range the union of \texttt{CardinalDirection} and \texttt{CentralPosition}. \texttt{CardinalDirection} is a finite enumeration of the values \texttt{East}, \texttt{North}, \texttt{NorthEast}, \texttt{NorthWest}, \texttt{South}, \texttt{SouthEast}, \texttt{SouthWest}, \texttt{West}. \texttt{CentralPosition} contains only the individual \texttt{Centre}. In addition, we introduced the defined classes \texttt{CentralArea}, \texttt{EastArea}, \texttt{NorthArea}, \texttt{NorthEastArea}, \texttt{NorthWestArea}, \texttt{WestArea}, \texttt{SouthEastArea}, \texttt{SouthWestArea}, \texttt{SouthArea}, whose subclasses, representing the buildings and gardens of the monastery, are deduced by inference. Figure \ref{imgNorthArea} shows the description of the class  \texttt{NorthArea}, while Figure \ref{imgNorthAreaInf}  illustrates the inferred hierarchy of  \texttt{NorthArea}.

In addition, we define the position of buildings or gardens in the map with respect to other contiguous buildings or gardens, by means of the object-properties \texttt{onEastOf}, \texttt{onNorthEastOf}, \texttt{onNorthOf}, \texttt{onNorthWestOf}, \texttt{onSouthEastOf}, \texttt{onSouthOf}, \texttt{onSouthWestOf}, \texttt{onWestOf}, where \texttt{onEastOf} is the inverse of \texttt{onWestOf}, \texttt{onNorthOf} of \texttt{onSouthOf}, \texttt{onNorthWestOf} of \texttt{onSouthEastOf}, and \texttt{onNorthEastOf} of \texttt{onSouthWestOf}.
\end{sloppypar}

\begin{figure}	
	\includegraphics[]{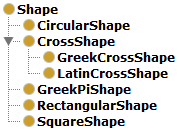}
	\caption{Subclass hierarchy of \texttt{Shape}.}
	\label{imgShape}
\end{figure}

Next we analyze the shape, the size, and the internal structure of buildings and gardens. 
\footnote{We take into account the information provided by our available sources, namely \cite{ Willis,DeVecchi95,Donofrio94,Gombrick1950,Kuback2001} and http://www.stgallplan.org/.} 
We define the class \texttt{Shape}, modeling the shape of structures, 
and whose subclass hierarchy is shown in Figure \ref{imgShape}, and the object-property \texttt{hasShape}. 
The class \texttt{Size} and the object-property \texttt{hasSize} model the size of buildings. 
Buildings in the map having the same size are associated to equivalent subclasses of the class \texttt{Size}.

 \begin{figure} [H]
	\includegraphics[scale=0.90]{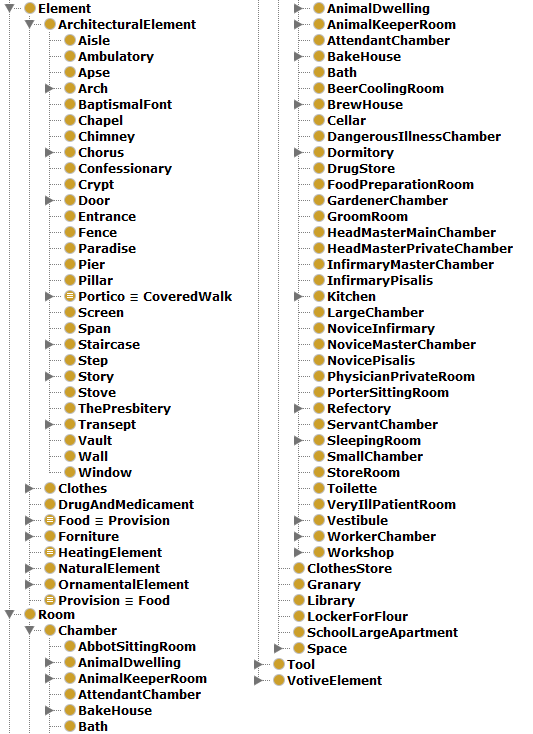}
	\caption{Subclass hierarchy of \texttt{Element}.}  
	\label{elementimg}
\end{figure}

\begin{figure}[H]	
	\includegraphics[]{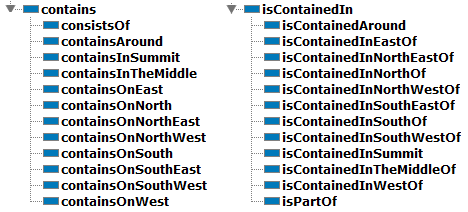}
	\caption{Object-properties related to the class \texttt{Element}.}
	\label{imgElement2}
\end{figure}

\begin{figure}	[H]
	\includegraphics[]{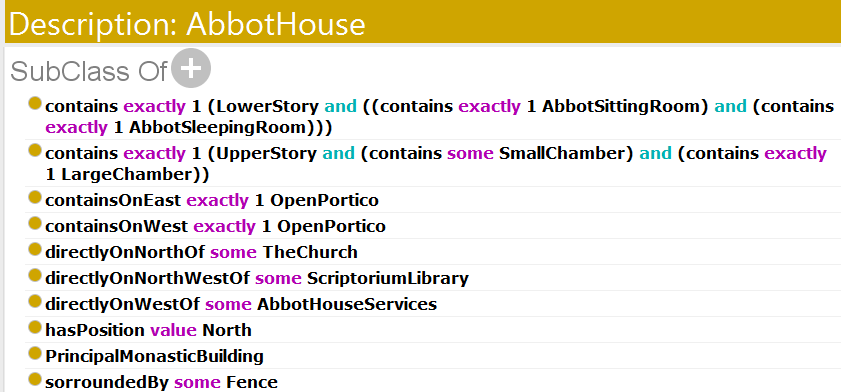}
	\caption{Description of \texttt{AbbotHouse}.}
	\label{imgAbbot}
\end{figure}
The class \texttt{Element} has as subclasses the class \texttt{ArchitecturalElement}, describing general architectural elements 
inside the map, the class \texttt{Forniture}, modeling objects used in  everyday life such as \texttt{Bedstand} and \texttt{Desk}, the class \texttt{Tool}, modeling tools of common use such as \texttt{Furnace} and \texttt{Boiler}, and classes describing rooms, clothes, food, votive, and ornamental elements.  
Anything included in such classes can be used in other contexts outside the Saint Gall plan.

   \begin{figure}[H]
       \centering	
   	\includegraphics[scale=0.8]{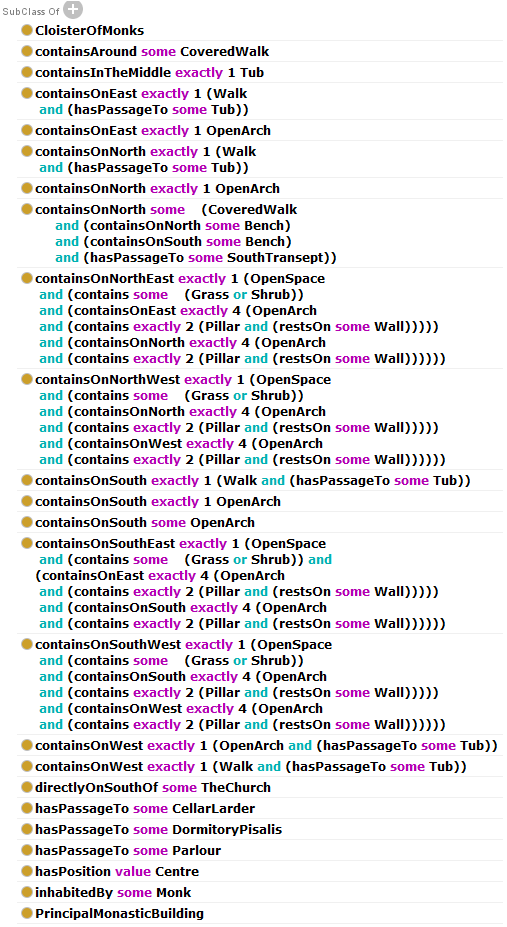}
   	\caption{The class \texttt{TheCloister}.}
   	\label{imgCloister}
   \end{figure}

In addition, we provide the object-properties \texttt{contains}, together with its subproperties \texttt{consistOf}, \texttt{containsAround}, and so on, and its inverse \texttt{isContainedIn}, together with its subproperties \texttt{isPartOf}, \texttt{isContainedAround}, and so on. The hierarchy of subclasses of \texttt{Element} and of their related properties are illustrated in Figure \ref{elementimg} and \ref{imgElement2}. 
In Figure \ref{imgAbbot} we show our representation of the abbot house.  

This building, inhabited by the abbot, is surrounded by a fence. It consists of two stories of which the lower one has an open portico on the east and west sides. The inner space is split into two chambers: the abbot sleeping and sitting rooms. The upper story contains some small chambers and one large chamber. Details concerning the furniture of the abbot sleeping and sitting rooms are modeled by the classes  \texttt{AbbotSleepingRoom} and \texttt{AbbotSittingRoom}, respectively, which are subclasses of \texttt{Chamber}. Our representation of the monk cloister can be found in Figure \ref{imgCloister}.

\begin{figure}[H]	
	\includegraphics[scale=0.6]{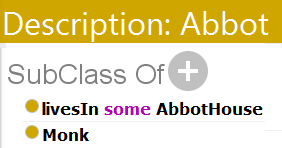}
	\caption{Description of  \texttt{Abbot}.}               
	\label{imgPerson}
\end{figure}

We also modeled people living in the monastery. As shown in Figure \ref{imgPerson}, they are classified according to the place in which they live and 
spend most of the day.  

\section{Conclusions}

We presented an ontology for the Saint Gall plan, that describes the
ideal model of the structure of a monastic Benedictine building. The plan allows one to study the most significant features of European monastic buildings such as the Monastery of San Nicol\`{o} l'Arena in Catania, the Abbey of Santa Giustina in Padua, the Cluny Abbey, the Montecassino Abbey \cite{DeVecchi95}.
By means of \textsf{SaintGall} Ontology, scholars and researchers in Human Science can effectively compare several distinct monastic architectures, and from their differences and similarities make inferences not only in the architectonic and stylistic ambits but also in the interpretative and theological areas \cite{Gombrick1950}.  


We are currently considering the integration of the \textsf{SaintGall} Ontology with the ontology for the Benedictine Monastery of Catania presented in \cite{JLIS} and other widespread ontologies for cultural heritage such as  as \textsf{CIDOC-CRM}.\footnote{http://www.cidoc-crm.org}
Some generic classes from the \textsf{SaintGall} Ontology, such as \texttt{Church} and \texttt{Cloister}, can be reused to design novel ontologies describing buildings outside the Benedectine context. Consider, for instance, the architectonic structure of closed garden (cloister or court), which can be also found in municipal buildings.  

The \textsf{SaintGall} Ontology was designed in such a way as to describe the SaintGall map in detail. That makes it more complex than both the ontology of the Monastery of Catania \cite{JLIS} and \textsf{Ontoceramic}
\cite{OntoCeramic}, an ontology for the classification of pottery. Moreover, the \textsf{SaintGall} Ontology cannot be represented in the set-theoretic fragment considered in \cite{CanNic13}, used in recent work by some of the authors for ontologies representation and reasoning. Thus, we intend to design a new decidable set-theoretic fragment admitting the composition operator allowing one to represent and reason on the \textsf{SaintGall} Ontology. Results in \cite{CanNicOrl10,CanNicOrl11}  are helpful to construct an appropriate decision procedure for such set-theoretic fragment.

\end{document}